\documentclass[letterpaper]{article} 
\usepackage{aaai25}  
\usepackage{times}  
\usepackage{helvet}  
\usepackage{courier}  
\usepackage[hyphens]{url}  
\usepackage{graphicx} 
\usepackage{natbib}  
\usepackage{caption} 
\usepackage{algorithm}
\usepackage{algorithmic}

\usepackage{algorithm}
\usepackage{algorithmic}
\usepackage{amsmath}
\usepackage{amssymb}
\usepackage{amsfonts}
\usepackage{amsthm}
\usepackage{booktabs}
\newtheorem{definition}{Definition}
\usepackage{multirow}
\usepackage{tabularx}
\usepackage{booktabs}
\usepackage{graphicx, subfigure}
\usepackage{bm}
\usepackage{url}
\usepackage{colortbl, xcolor}
\usepackage{tikz}

\usepackage{newfloat}
\usepackage{listings}
\DeclareCaptionStyle{ruled}{labelfont=normalfont,labelsep=colon,strut=off} 
\floatstyle{ruled}
\newfloat{listing}{tb}{lst}{}
\floatname{listing}{Listing}

\title{Community-Centric Graph Unlearning}
\author{
    Yi Li\textsuperscript{\rm 1,2},
    Shichao Zhang\textsuperscript{\rm 1,2},
    Guixian Zhang\textsuperscript{\rm 3},
    Debo Cheng\textsuperscript{\rm 4}\thanks{Corresponding author}
}
\affiliations{
    \textsuperscript{\rm 1}Key Lab of Education Blockchain and Intelligent Technology, Ministry of Education,\\ Guangxi Normal University, Guilin, 541004, China\\
    \textsuperscript{\rm 2}Guangxi Key Lab of Multi-Source Information Mining and Security,\\ Guangxi Normal University, Guilin, 541004, China\\
    \textsuperscript{\rm 3}School of Computer Science and Technology, China University of Mining and Technology,\\ Xuzhou, Jiangsu, 221116, China\\
    \textsuperscript{\rm 4}UniSA STEM, University of South Australia,\\ Mawson Lakes, Adelaide, Australia\\
    liiyi.xsjl@gmail.com, zhangsc@mailbox.gxnu.edu.cn, guixian@cumt.edu.cn, chedy055@mymail.unisa.edu.au
}

\usepackage{bibentry}

\begin{document}

\maketitle

\begin{abstract}
Graph unlearning technology has become increasingly important since the advent of the `right to be forgotten' and the growing concerns about the privacy and security of artificial intelligence. Graph unlearning aims to quickly eliminate the effects of specific data on graph neural networks (GNNs). However, most existing deterministic graph unlearning frameworks follow a balanced partition-submodel training-aggregation paradigm, resulting in a lack of structural information between subgraph neighborhoods and redundant unlearning parameter calculations. To address this issue, we propose a novel \underline{G}raph \underline{S}tructure \underline{M}apping \underline{U}nlearning paradigm (\textbf{GSMU}) and a novel method based on it named \underline{\textbf{C}}ommunity-centric \underline{\textbf{G}}raph \underline{\textbf{E}}raser (\textbf{CGE}). CGE maps community subgraphs to nodes, thereby enabling the reconstruction of a node-level unlearning operation within a reduced mapped graph. CGE makes the exponential reduction of both the amount of training data and the number of unlearning parameters. Extensive experiments conducted on five real-world datasets and three widely used GNN backbones have verified the high performance and efficiency of our CGE method, highlighting its potential in the field of graph unlearning. \url{https://github.com/liiiyi/CCGU}
\end{abstract}

%

\section{Introduction}

As Artificial Intelligence (AI) permeates various sectors, protecting individual privacy and data security has become an increasingly pressing concern~\cite{fux}. Legal frameworks, such as the ``right to be forgotten"~\cite{law1,law2,law3},  alongside the rise of techniques like membership inference attacks~\cite{mia} have underscored the need for model providers to swiftly remove specific data and mitigate its impacts. In parallel, Graph Neural Networks (GNNs)~\cite{ccgnn,gin,lic,liz} have gained prominence as a powerful tool,  renowned for their ability to model complex data relationships. However, their widespread application in areas involving highly sensitive data~\cite{guix}, such as social networks\cite{smedia} and recommendation systems~\cite{rec}, has intensified the demand for effective privacy-preserving techniques. Among these, \textbf{\textit{Graph Unlearning}}~\cite{grapheraser} has gained particular prominence. 

Graph unlearning seeks to precisely quantify and remove the influence of specific data within graph-structured networks while maintaining the integrity of the overall model, thereby addressing critical privacy concerns in an era of expanding AI applications. A straightforward approach to unlearning involves retraining the entire model from scratch. However, this method is highly impractical due to the substantial time costs and associated model downtime. Consequently, current research in unlearning has focused on developing techniques that reduce this time expenditure while maintaining the model's effectiveness~\cite{gusurvey}. 

To achieve efficient and effective unlearning, researchers have developed various strategies that circumvent the costly process of full retraining~\cite{approxi2,approxi3,sisa,grapheraser}.
These unlearning frameworks are typically divided into approximate and deterministic methods~\cite{musurvey}. Approximate methods~\cite{approxi2,approxi3} involve fine-tuning models to implement unlearning; however, they only provide abstract statistical guarantees of privacy. In contrast, deterministic methods, the focus of this paper, involve retrospective retraining to remove data effects entirely~\cite{arcane}. For example, the SISA~\cite{sisa} approach partitions data for selective retraining. GraphEraser~\cite{grapheraser}, building on SISA, uses balanced partitioning to ensure equal node distribution and aggregates submodels using a learning-based method. Recently introduced, GUIDE~\cite{guide} enhances predictive performance by repairing edges before aggregating partitioned subgraphs.

In summary, the prevailing deterministic graph unlearning frameworks~\cite{grapheraser,guide,receraser} are primarily based on the Balance Partition-Submodel Training-Aggregation paradigm (\textbf{BP-SM-TA}). The principal limitation of this paradigm lies in its assumption that the unlearning operation can be decomposed into a set of discrete sub-problems, each with its own global representativeness. This decomposition, however, sacrifices the connectivity between sub-problems and increases the complexity of managing them. Graph unlearning with BP-SM-TA, two specific challenges arise:
 
\textbf{\textit{Challenge 1} (Low Structural Utilization)}: In the BP-SM-TA paradigm, submodels are treated as independent, leading to the disassembly of the original graph during partitioning and resulting in the loss of critical structural information between shards that were originally connected. This, in turn, diminishes the overall comprehension of the graph's structure. Additionally, the balanced partitioning method, which aims to achieve uniform training times for submodels by selecting nodes through clustering, further compromises the graph's structural integrity.

\textbf{\textit{Challenge 2} (High Model Complexity)}: In the BP-SM-TA paradigm, submodels must independently represent the original graph, meaning each submodel needs to capture the essential features of the original structure. This requirement introduces significant challenges in terms of partitioning and structuring the submodels. During the unlearning process, this paradigm necessitates retraining all affected submodels, which leads to extensive parameter computations. For large graphs, simultaneously loading all submodels for aggregation incurs substantial time costs, which contradicts the primary objective of efficient graph unlearning.

To tackle the aforementioned challenges, we propose a novel \underline{G}raph \underline{S}tructure \underline{M}apping \underline{U}nlearning paradigm (\textbf{GSMU}). Specifically, GSMU generates a mapped graph comprising the non-redundant features and structural information of the original graph. The nodes and edges of the mapped graph are derived from the subgraphs and inter-subgraph node relationships in the original graph. The construction of edges between mapped nodes addresses the limitation of submodel independence identified in \textit{Challenge 1}. Meanwhile, unlearning operations only require updating the affected mapped nodes and edges, thereby avoiding extensive parameter retraining and addressing the limitation of independent submodel representativeness identified in \textit{Challenge 2}. 
Moreover, we propose a novel community-centric graph unlearning method, \underline{C}ommunity-centric \underline{G}raph \underline{E}raser (\textbf{CGE}), which serves as a practical implementation of GSMU. CGE comprises two components: parameter-free community-centric graph mapping and a node-level unlearning strategy. Specifically, CGE divides the original graph into multiple communities to accomplish the subgraph-node mapping required by the GSMU, and subsequently the unlearning strategy only needs to be performed on the nodes after the mapping, thus controlling the part affected by the unlearning requirement at the node level. Our contributions are summarized as follows:

\begin{itemize}
    \item We propose a novel graph structure mapping unlearning paradigm, \textbf{GSMU}, to overcome the space and time constraints of the traditional BP-SM-TA paradigm in addressing unlearning concerns. To the best of our knowledge, this is the first application of a mapped graph approach in graph unlearning.
    \item Based on GSMU, we introduce a new efficient graph unlearning framework, \textbf{CGE}, which supports deterministic data unlearning while maximizing the retention of the original graph's structural information and semantic feature. By mapping communities to nodes, unlearning operations can be executed at the node level.
    \item Extensive experiments on five real-world datasets demonstrate that our proposed CGE achieves superior performance and remarkable unlearning efficiency across datasets of varying scales and characteristics.
\end{itemize}

\section{Preliminaries}

\paragraph{Notations.} In this paper, we use specific notations to represent various components of GNNs and related operations. The tilde notation $\widetilde{\cdot}$, is used to indicate entities produced by mapping operations. Calligraphic fonts are employed to denote sets. We summarize the notations in Table \ref{notations}.

\begin{table}[t]\small

\begin{center}
    \begin{tabular}{ l l }
        \toprule[1.5pt]
        \textbf{Notation} & \textbf{Description} \\
        \midrule[1pt]
        $\mathcal{G}=(\mathcal{V}, \mathcal{E})$ & Original Graph \\
        $\mathcal{\widetilde{G}}=(\mathcal{\widetilde{V}}, \mathcal{\widetilde{E}})$ & Mapped Graph \\
        $\mathcal{G} \mapsto \mathcal{\widetilde{G}}$ & The Mapping from $\mathcal{G}$ to $\mathcal{\widetilde{G}}$ \\
        $\mathcal{C} = \{C_1, C_2, \ldots, C_k\}$ & Node Subgraphs Set \\
        $\mathcal{V}_u = \{v_{u1}, v_{u2}, \ldots v_{un} \}$ & Unlearning Requests Set \\
        $\mathcal{I}_{\mathcal{V}_u}$ & Unlearning Influence Set \\
        $\operatorname{Update_\mathcal{\widetilde{G}}}(\widetilde{\mathcal{V}}', \widetilde{\mathcal{E}}', \widetilde{\mathcal{X}}', \widetilde{\mathcal{Y}}')$ & The process of updating $\mathcal{\widetilde{G}}$ \\
        \bottomrule[1.5pt]
    \end{tabular}
    \caption{Summary of Notations.}
    \label{notations}
\end{center}
\end{table}

\paragraph{Deterministic Graph Unlearning.} Given a graph $\mathcal{G} = (\mathcal{V}, \mathcal{E})$, where $\mathcal{V}$ represents the set of nodes and $\mathcal{E}$ represents the set of edges, and a GNN backbone model denoted as $f_{\theta}(\mathcal{G})$, with $\theta$ being the model parameters. Suppose the unlearning request is a sequence of nodes $\mathcal{V}_u \subseteq \mathcal{V}$. The goal of deterministic graph unlearning is to eliminate the influence of these nodes from the model, formalized as:
\begin{equation}
    f_{\theta'}(\mathcal{G}') \approx f_{\theta}(\mathcal{G}) \quad \text{subject to} \quad \mathcal{V}_u \notin \mathcal{G}',
\end{equation}
where $\mathcal{G}' = (\mathcal{V} \setminus \mathcal{V}_u, \mathcal{E} \setminus E_u)$ and $\theta'$ denotes the updated model parameters after unlearning. The advanced objectives of graph unlearning should ensure portability, comparable model utility to training from \textit{Scratch}~\cite{grapheraser}, the maintenance of model performance before and after unlearning, high graph structure utilization, and minimal unlearning window periods. Graph unlearning can be broadly divided into node-level and edge-level approaches. In our GSMU paradigm, edge unlearning is considered a subproblem of node unlearning. Consequently, our work focuses on the node classification task, treating node unlearning as the fundamental unit.

\paragraph{Community Detection.} Community detection aims to partition a graph $\mathcal{G}=(\mathcal{V}, \mathcal{E})$ into a community set $\mathcal{C} = \{C_1, C_2,\ldots, C_k\}$, optimizing intra-community connectivity while minimizing inter-community connectivity~\cite{cdsurvey1,cdsurvey2}. Hierarchical community detection~\cite{oslom} further refines this process by creating multi-level community structures, which clarify relationships at each level. In this paper, we first perform coarse-grained segmentation followed by fine-grained segmentation, as described in the next section.

\begin{figure*}[t]
\centering
\includegraphics[width=0.9\textwidth]{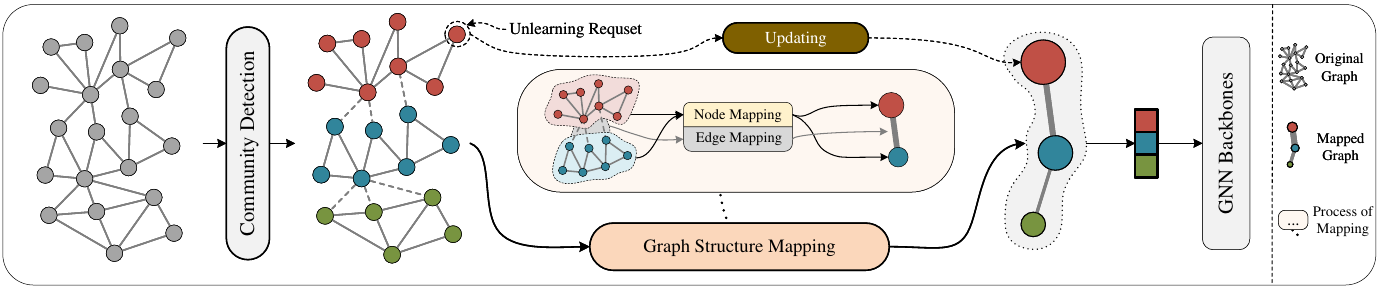} 
\caption{Community-centric Graph Eraser (CGE) framework. CGE maps the communities in original graph $\mathcal{G}$ to obtain graph $\mathcal{\widetilde{G}}$ by graph structure mapping. Only the related mapped nodes need to be updated when nodes require unlearning.}
\label{fig:framework}
\end{figure*}

\section{Community-Centric Unlearning}
In this section, we detail the GSMU paradigm and the specific implementation of CGE.

\paragraph{GSMU.} In contrast to the BP-SM-TA paradigm, GSMU employs a subgraph-based graph structure mapping approach to derive a representative mapped graph, $\mathcal{\widetilde{G}}$, from the original graph, $\mathcal{G}$. Each mapped node in $\mathcal{\widetilde{G}}$ corresponds to a subgraph in $\mathcal{G}$. After partitioning, $\mathcal{G}$ and $\mathcal{\widetilde{G}}$ are isolated to ensure privacy. When it is necessary to unlearn a node, the unlearning operation only affects its corresponding mapped nodes, i.e., the subgraphs in which the node resides.

\begin{definition}
\label{def1}
    \textbf{Graph Structure Mapping} Given a graph $\mathcal{G} = (\mathcal{V}, \mathcal{E})$ with a set of predefined subgraphs $\mathcal{C}$, GSMU maps it to a graph $\mathcal{\widetilde{G}} = (\mathcal{\widetilde{V}}, \mathcal{\widetilde{E}})$. Each subgraph $C \in \mathcal{C}$ constructs a mapped node, i.e., $\mathcal{V} \mapsto \mathcal{\widetilde{V}} = \{ \widetilde{v}_i \mid C \in \mathcal{C} \}$. An edge $\widetilde {e}_{ij}$ is created for each edge $e_{ij} \in \mathcal{E}$ that connects nodes in different subgraphs $C_i$ and $C_j$, i.e., $\mathcal{E} \mapsto \mathcal{\widetilde{E}} = \{ \widetilde{e}_{ij} \mid e_{ij} \in \mathcal{E}, \; e_{ij} \text{ connects } C_i \text{ and } C_j \}$. Node features fusion is performed on each subgraph, i.e., $\mathcal{X} \mapsto \mathcal{\widetilde{X}} = \{ \widetilde{\mathbf{x}}_i \mid \widetilde{\mathbf{x}}_i = f(\{\mathbf{x}_v \mid v \in \mathcal{V}_i\}) \}$, where $f$ is a fusion function. Node labels are determined by voting within each subgraph, i.e., $\mathcal{Y} \mapsto \mathcal{\widetilde{Y}} = \{ \widetilde{y}_i \mid \widetilde{y}_i = g(\{y_v \mid v \in \mathcal{V}_i\}) \}$, where $g$ is a labeling function. \textbf{The mapping relationship between $\mathcal{G}$ and $\widetilde{\mathcal{G}}$ is defined as: $\mathcal{G} \mapsto \widetilde{\mathcal{G}} = \left\{ \mathcal{V} \mapsto \mathcal{\widetilde{V}}, \; \mathcal{E} \mapsto \mathcal{\widetilde{E}}, \mathcal{X} \mapsto \mathcal{\widetilde{X}}, \; \mathcal{Y} \mapsto \mathcal{\widetilde{Y}} \right\}$}.
\end{definition}

Under the developed GSMU framework, we propose a community-centric graph unlearning method, i.e., \underline{C}ommunity-Centric \underline{G}raph \underline{E}raser (CGE). The subsequent parts will provide an overview of the operational process of CGE and explain how CGE constructs mapped graph and utilises if for node prediction and unlearning.

\subsection{Overview of CGE}
The CGE framework, illustrated in Figure 1, consists of two stages: the generation of a community-centered mapped graph and a node-level unlearning/training stage. In particular, CGE employs hierarchical community detection to construct a subgraph set $\mathcal{C}$, establish the graph structure mapping in Definition \ref{def1}, and generate a mapped graph $\mathcal{\widetilde{G}}$ for subsequent operations.

\begin{itemize}
    \item \textbf{Unlearning:} Upon receiving a sequence of unlearning requests $\mathcal{V}_u$, CGE employs the node-level unlearning strategy to determine the specific influence $\mathcal{I}_{\mathcal{V}_u}$ of $\mathcal{V}_u$ on the mapped graph $\mathcal{\widetilde{G}}$ identifying the specific affected mapped edges and nodes. Subsequently, CGE recalculates the nodes and edges within $\mathcal{I}_{\mathcal{V}_u}$ according to the mapping, thereby updating the graph $\mathcal{\widetilde{G}}$.
    \item \textbf{Prediction:} When the node $v_p$ requires prediction, CGE determines destination community $C_i$ based on the community of the adjacent nodes of $v_p$. The node features of $v_p$ are then fused with the mapped nodes $\widetilde{v}_i$ corresponding to $C_i$, which in turn updates the graph $\mathcal{\widetilde{G}}$ acting on the GNN. The node $\widetilde{v}_i$ is reanalyzed to map the attributes of the node $v_p$.
\end{itemize}

\subsection{Community-Centric Mapping}
CGE generates a community-centered mapped graph $\widetilde{\mathcal{G}}$ and establishes a graph structure mapping $\mathcal{G} \mapsto \widetilde{\mathcal{G}}$, as described in Definition \ref{def1}. In particular, $\widetilde{\mathcal{G}}$ is utilized solely to retain the representative information of $\mathcal{G}$, thereby enabling its replacement for representation learning. Each community within $\mathcal{G}$ is mapped into a node, and the robustness score between the mapped node pairs is employed to construct the edge relationships. 

\subsubsection{Community Detection}
CGE introduces the Louvain~\cite{louvain} to initialize communities and uses the OSLOM method~\cite{oslom} to optimize community structure. Louvain is a community detection algorithm based on modularity optimization. It attempts to move each node into the community of its neighboring nodes to maximize the modularity increment.  The modularity is calculated as follows:
\begin{equation}
Q = \frac{1}{2m} \sum_{ij} \left[ A_{ij} - \frac{k_i k_j}{2m} \right] \delta(c_i, c_j),
\end{equation}
where $A_{ij}$ represents the edge weight between node $v_i$ and node $v_j$. Additionally, $k_i$ and $k_j$ denote the degrees of $v_i$ and $v_j$, respectively. The term $m$ denotes the total weight of all edges in $\mathcal{G}$. $\delta(c_i, c_j)$ is an indicator variable that equals 1 when nodes $i$ and $j$ belong to the same community, and 0 otherwise.

OSLOM calculates the statistical significance of the initialized communities using the following formula:
\begin{equation}
p(C) = \sum_{t \geq t_C} \binom{T_C}{t} p^t (1-p)^{T_C-t},
\end{equation}
where $T_C$ represents the total number of possible connections in community $C$, $t_C$ represents the actual number of connections in $C$, and $p$ represents the probability that nodes in $C$ are connected to other nodes. OSLOM then returns an optimized community set $\mathcal{C} = \{C_1, C_2, \ldots, C_k\}$.

\subsubsection{Mapping}
As delineated in Definition \ref{def1}, the mapping of $\mathcal{G} \mapsto \widetilde{\mathcal{G}}$ comprises four distinct mappings. Among these, we categorize the node, feature, and label mapping as \textbf{node mapping}, which align with \textbf{edge mapping}.

\paragraph{Node Mapping.}

The initial step is to construct a node mapping based on the community set $\mathcal{C}$:
\begin{equation}
    \mathcal{V} \longmapsto \mathcal{\widetilde{V}} = \{ \widetilde{v}_i \mid C \in \mathcal{C} \}.
\end{equation}

To ensure that supervised learning goes well, it is essential to assign the correct labels and features to the mapped nodes. Despite the strict division of communities, node features within a community may vary and have inconsistent labels, making it impractical to average the node labels and features simply~\cite{zhangknn}. To address this, we calculated the core feature components of the community and designed a feature-based label mapping method. We construct a feature matrix $\mathbf{X}_{C} \in \mathbb{R}^{n \times d}$, where $d$ is the feature space dimensionality and rows correspond to node feature vectors in the community $C$. Principal Component Analysis (PCA) is then applied to $\mathbf{X}_{C}$ to reduce its dimensionality. Let $\mathbf{X}_{C}^\prime \in \mathbb{R}^{n \times k}$ be the transformed feature matrix by PCA, where $k$ is the number of principal components based on the ratio $n$ components. The community feature $\mathbf{\widetilde{f}}_{C}$ is calculated as the mean of the transformed feature vectors:
\begin{equation}
    \widetilde{\mathbf{f}}_{C} = \frac{1}{|\mathcal{V}_C|} \sum_{v \in \mathcal{V}_C} \mathbf{x}_v',
\end{equation}
where $\mathbf{x}_v'$ is the transformed feature vector of node $v$ in community $C$. The feature mapping is then constructed using the fusion feature:
\begin{equation}
    \mathcal{X} \longmapsto \mathcal{\widetilde{X}} = \{ \widetilde{\mathbf{x}}_i \mid \widetilde{\mathbf{x}}_i = \widetilde{\mathbf{f}}_{C_i} \text{ for } C_i \in \mathcal{C} \},
\end{equation}
where $\widetilde{\mathbf{f}}_{C_i}$ represents the fusion feature of community $C_i$.

Next, we identify a subset of nodes within the current community $C$ that have the highest feature similarity to the mapped node $\widetilde{v}$, and then perform majority voting on these nodes. These nodes are then subjected to a majority voting process. Specially, we consider a community $C$ with node labels $\{l_1, l_2, \ldots, l_n\}$ and their corresponding feature vectors $\mathbf{X}_{C}$. We calculate the Euclidean distance between each node feature vector $x_i \in \mathbf{X}_{C}$ in the community $C$ and $\mathbf{\widetilde{f}}_{\mathcal{C}}$.
\begin{equation}
    d_i = \|\widetilde{\mathbf{f}}_{C} - \mathbf{x}_{i}\|_2.
\end{equation}
We sort the distances in ascending order and calculate the gradient $g_i = d_{i+1} - d_i$ between consecutive distances. The distance corresponding to the maximum gradient is given by:
\begin{equation}
    \tau = d_{\arg\max(g)}.
\end{equation}
Finally, we select nodes with distances $d_i \leq \tau$ and aggregate their labels using majority voting:
\begin{equation}
    \widetilde{l}_{C} = \text{mode}(\{l_i \mid d_i \leq \tau\}).
\end{equation}

The label mapping is then constructed as:
\begin{equation}
    \mathcal{Y} \longmapsto \mathcal{\widetilde{Y}} = \{ \widetilde{y}_i \mid \widetilde{y}_i = \widetilde{l}_{C_i} \text{ for } C_i \in \mathcal{C} \}.
\end{equation}

\paragraph{Edge Mapping.}
Using the adjacency matrix $\mathbf{A}$ to represent the graph $\mathcal{G}$, for each edge $(u, v) \in \mathcal{E}$ where node $u$ belongs to community $C_i$ and node $v$ belongs to community $C_j$, we define the edge count between the community pair $(C_i, C_j)$ as $s_{ij} = \sum_{u \in C_i} \sum_{v \in C_j} \mathbf{A}_{uv}$. The robustness score $R_{ij}$ of the connection between communities $C_i$ and $C_j$ is evaluated using the following formula:
 
\begin{equation}
    R_{ij} = (\frac{s_{ij}}{\sqrt{D_i^{\text{out}}}}) \times (\frac{s_{ij}}{\sqrt{D_j^{\text{in}}}}) + \frac{s_{ij}}{|C_i \cup C_j|},
\end{equation}
where $D_i^{\text{out}}$ and $D_j^{\text{in}}$ denote the out-degree of $C_i$ and the in-degree of $C_j$, respectively. 

A smoothing function, $w_{ij} = \lambda \times \exp(-R_{ij}) + \eta$ is then employed to process the robustness score, representing the edge weight of the original community relationship. To store these weights, we introduce a set $\mathcal{W}$, where each element is a key-value pair: the key is the node pair $(i, j)$, and the value is $w_{ij}$. Subsequently, a threshold $\sigma$ is utilized to filter and construct the edge mapping $\mathcal{E} \rightarrow \widetilde{\mathcal{E}}$ as follows:
\begin{equation}
    \mathcal{E} \longmapsto \widetilde{\mathcal{E}} = \{ w_{ij} \mid (C_i, C_j) \in \mathcal{W} \text{ and } w_{ij} \geq \sigma \}.
\end{equation}

Finally, the total mapping relationship is established as follows:
\begin{equation}
    \mathcal{G} \longmapsto \widetilde{\mathcal{G}} = \left\{ \mathcal{V} \mapsto \mathcal{\widetilde{V}}, \; \mathcal{E} \mapsto \mathcal{\widetilde{E}}, \mathcal{X} \mapsto \mathcal{\widetilde{X}}, \; \mathcal{Y} \mapsto \mathcal{\widetilde{Y}} \right\}.
\end{equation}

\subsection{Node-Level Unlearning Strategy}

Upon receipt of an unlearning request as a sequence, denoted as $\mathcal{V}_u = \{v_{u1}, v_{u2}, \ldots v_{un} \}$, CGE initially identifies the communities of the nodes in $\mathcal{V}_u \subset \mathcal{V}$ for this batch and subsequently returns a set of affected communities, denoted as $\mathcal{C}_u$. Based on the graph structure mapping $\mathcal{G} \mapsto \widetilde{\mathcal{G}}$, the set of all affected mapped nodes and edges, denoted as $\mathcal{I}_{\mathcal{V}_u}$, is obtained. 

It is important to note that the application of node-level PCA computation acts as a \textit{filter} to exclude non-principal components and unlearning requests that are not affected. If a node $v_{ui}$ is not utilized in the calculation of the features and labels associated with the mapped node, it can be inferred that the information contained within the node is not retained within the mapped graph. Consequently, the influence of node-level unlearning of $v_{ui}$ is effectively negated, thereby avoiding unnecessary expenditure of time and resources.

Based on $\mathcal{I}_{\mathcal{V}_u}$, CGE updates the relevant feature and structure information of the graph and generates an updated graph $\mathcal{\widetilde{G}}'$:

\begin{equation}
    \widetilde{\mathcal{G}}' = \operatorname{Update_\mathcal{\widetilde{G}}}(\widetilde{\mathcal{V}}', \widetilde{\mathcal{E}}', \widetilde{\mathcal{X}}', \widetilde{\mathcal{Y}}').
\end{equation}
  
The updated graph $\mathcal{\widetilde{G}}'$ and the mapped graph $\mathcal{\widetilde{G}}$ must satisfy the following constraints to completely eliminate specific data and its derived information within the model.
\begin{equation}
    f_{\theta'}(\widetilde{\mathcal{G}}') \approx f_{\theta}(\mathcal{\widetilde{G}}) \quad \text{subject to} \quad \mathcal{V}_u \notin \widetilde{\mathcal{G}}'.
\end{equation}

\section{Experiments}
To evaluate the efficacy of CGE, we conducted a series of comprehensive experiments using four real-world datasets and three prevalent GNN backbones. The evaluation addresses the following questions: (1) Can CGE provide excellent and comparable model utility? (2) How efficient is CGE in practical applications? (3) Can CGE achieve deterministic graph unlearning? Additionally, a series of ablation studies were performed to examine CGE's superiority at each stage of the graph unlearning process. We also assessed the applicability of different community detection methods to CGE, as detailed in Appendix D.

\begin{table}[t]

\begin{center}
\small
    \begin{tabularx}{\columnwidth}{lXXcXX}
        \toprule[1.5pt]
        \textbf{Dataset} & \textbf{\#Nodes} & \textbf{\#Edges} & \textbf{\#Classes} & \textbf{\#Features} & \textbf{Density} \\
        \midrule[1pt]
        Cora & 2,708 & 5,429 & 7 & 14,33 & 0.148\% \\
        Citeseer & 3,327 & 4,732 & 6 & 3,703 & 0.085\% \\
        CS & 18,333 & 163,788 & 15 & 6,805 & 0.097\% \\
        Reddit & 232,965 & \text{$1.1\times10^8$} & 41 & 602 & 0.040\% \\
        \bottomrule[1.5pt]
    \end{tabularx}
    \caption{Statistics of Datasets.}
    \label{datasets}
\end{center}
\end{table}

\subsection{Experimental Setup}

\subsubsection{Datasets.} We evaluated the CGE on four real-world datasets of various sizes, including Cora~\cite{cc}, Citeseer~\cite{cc}, CS~\cite{cs} and Reddit \footnote{\url{https://docs.dgl.ai/generated/dgl.data.RedditDataset.html}}, all of which are commonly used in GNN evaluations. The large-scale Reddit dataset was specifically included to assess the unlearning framework's performance in a real-world context. The statistics for these datasets are summarized in Table \ref{datasets}.

\subsubsection{GNN Backbones.} We equipped our CGE with three GNN backbones, GCN~\cite{gcn}, GAT~\cite{gat}, and GraphSAGE~\cite{sage}, to evaluate its versatility. Detailed descriptions of the three backbones can be found in Appendix A.

\subsubsection{Baselines.} To demonstrate the efficacy of CGE, we compare it with the state-of-the-art unlearning algorithms, including SISA~\cite{sisa}, GraphEraser~\cite{grapheraser}, and GUIDE~\cite{guide}, all of which are deterministic unlearning methods based on the BP-SM-TA paradigm.  We also introduced \textit{Scratch}, a scheme involving retraining from beginning to end, to evaluate comparable model utility. For each unlearning batch, 0.5\% of the original dataset's nodes were randomly selected. More detailed experimental settings are provided in Appendix B.

\subsubsection{Metrics.} We consider two key aspects to measure the performance of CGE: model utility and unlearning efficiency. Model utility is assessed using the \textit{Macro F1 score}, while unlearning efficiency is evaluated based on the \textit{time cost} required for model deployment and unlearning.

\subsubsection{Implementation.}  CGE is implemented using Python 3.8.19 and DGL\footnote{\url{https://www.dgl.ai}}.  All experiments were conducted on an NVIDIA Tesla A800 GPU server running the Ubuntu 23.04 LTS operating system. 

\begin{table*}[htbp]\small
\centering

\newcolumntype{Z}{>{\centering\let\newline\\\arraybackslash\hspace{0pt}}X}
\begin{tabularx}{\textwidth}{l l >{\columncolor[gray]{0.95}}Z ZZZZZ}
    \toprule[1.5pt]
        \multirow{2}{*}{\textbf{Dataset}} & \multirow{2}{*}{\textbf{Model}} & \multicolumn{5}{c}{\textbf{Method}} \\
        \cmidrule(lr){3-7}
        & & \textbf{\textit{Scratch}} & \textbf{SISA} & \textbf{GraphEraser} & \textbf{GUIDE} & \textbf{CGE} \\
        \midrule
        \multirow{3}{*}{\textbf{Cora}} & SAGE & $0.7638 \pm 0.0018$ & $0.6024 \pm 0.0073$ & $0.6910 \pm 0.0032$ &  \underline{$0.7226 \pm 0.0031$} & $\bm{0.8745 \pm 0.0043}$ \\
                          & GAT & $0.7435 \pm 0.0039$ & $0.5793 \pm 0.0027$ & \underline{$0.7324 \pm 0.0051$} & $0.6640 \pm 0.0049$ & $\mathbf{0.7463 \pm 0.0083}$ \\
                          & GCN & $0.7675 \pm 0.0012$ & $0.5190 \pm 0.0057$ & $0.6992 \pm 0.0009$ & \underline{$0.7403 \pm 0.0093$} & $\mathbf{0.7586 \pm 0.0092}$ \\
        \hline
        \multirow{3}{*}{\textbf{Citeseer}} & SAGE & $0.7232 \pm 0.0003$ & $0.5189 \pm 0.0021$ &$0.7112 \pm 0.0022$ & $\mathbf{0.7271 \pm 0.0022}$ & \underline{$0.7170 \pm 0.0103$} \\
                              & GAT & $0.7194 \pm 0.0006$ & $0.5020 \pm 0.0007$ & $0.7004 \pm 0.0041$ & \underline{$0.7102 \pm 0.0019$} & $\mathbf{0.7473 \pm 0.0059}$ \\
                              & GCN & $0.7197 \pm 0.0013$ & $0.4771 \pm 0.0011$ & $0.6983 \pm 0.0062$ & \underline{$0.7033 \pm 0.0011$} & $\mathbf{0.7082 \pm 0.0003}$ \\
        \hline
        \multirow{3}{*}{\textbf{CS}} & SAGE & $0.8913 \pm 0.0007$ & $0.5920 \pm 0.0004$ & $0.7033 \pm 0.0007$ & \underline{$0.8016 \pm 0.0013$} & $\mathbf{0.8460 \pm 0.0004}$ \\
                            & GAT & $0.8811 \pm 0.0014$ & $0.6611 \pm 0.0011$ & \underline{$0.7227 \pm 0.0014$} & $0.6939 \pm 0.0009$ & $\mathbf{0.7664 \pm 0.0039}$ \\
                            & GCN & $0.8907 \pm 0.0023$ & $0.7004 \pm 0.0007$ & $0.6313 \pm 0.0022$ & \underline{$0.7053 \pm 0.0007$} & $\mathbf{0.7806 \pm 0.0032}$ \\
        \hline
        \multirow{3}{*}{\textbf{Reddit}} & SAGE & $0.9443 \pm 0.0024$ & $0.5901 \pm 0.0013$ & $0.5793 \pm 0.0085$ & \underline{$0.8201 \pm 0.0012$} & $\mathbf{0.9451 \pm 0.0015}$ \\
                            & GAT & $0.9477 \pm 0.0018$ & $0.6611 \pm 0.0029$ & $0.7865 \pm 0.0029$ & \underline{$0.8817 \pm 0.0091$} & $\mathbf{0.9138 \pm 0.0006}$ \\
                            & GCN & $0.9312 \pm 0.0061$ & $0.6882 \pm 0.0044$ & $0.7727 \pm 0.0008$ & \underline{$0.8033 \pm 0.0009$} & $\mathbf{0.9302 \pm 0.0011}$ \\
        \bottomrule[1.5pt]
\end{tabularx}
\caption{The comparison of Macro F1 scores across different graph unlearning frameworks on four datasets and three GNN backbones is shown. \textit{Scratch} is a retraining scheme from beginning to end, which is used to evaluate the utility of comparable models. The best-performing unlearning framework's performance is in \textbf{bold}, and the runner-up's performance is \underline{underlined}.}
\label{tab:metrics}
\end{table*}

\begin{table*}[htbp]\small
\centering

\newcolumntype{Z}{>{\centering\let\newline\\\arraybackslash\hspace{0pt}}X}
\begin{tabularx}{\textwidth}{lZZZZZZZZ}
    \toprule[1.5pt]
    \multirow{2}{*}{\textbf{Method}} & \multicolumn{4}{c}{\textbf{Model Deployment}} & \multicolumn{4}{c}{\textbf{Unlearning}}\\
    \cmidrule(lr){2-5}
    \cmidrule(lr){6-9}
    & \textbf{Cora} & \textbf{Citeseer} & \textbf{CS} & \textbf{Reddit} & \textbf{Cora} & \textbf{Citeseer} & \textbf{CS} & \textbf{Reddit} \\
    \midrule
    \textbf{\textit{Scratch}} & 131.52 & 109.62 & 702.11 & 13048.67 & 130.23 & 110.05 & 700.12 & 13056.33 \\
    \cmidrule(lr){1-9}
    \textbf{GraphEraser} & 179.53 & 167.16 & 1042.07 & 61220.05 & 10.78 & 12.13 & 62.56 & 3128.22\\
    \textbf{GUIDE} & 124.58 & 127.76 & \textbf{745.10} & 39064.71 & 10.11 & 11.07 & 71.60 & 3411.89\\
    \textbf{CGE} & \textbf{93.12} & \textbf{108.35} & 747.82 & \textbf{18143.95} & \textbf{8.12} & \textbf{9.58} & \textbf{9.33} & \textbf{49.52}\\ 
    \bottomrule[1.5pt]
\end{tabularx}
\caption{The comparison of two-stage time consumption of different graph unlearning frameworks on four datasets (second).}
\label{tab:eff}
\end{table*}

\subsection{Model Utility of CGE}
To answer question (1), we conducted a comparative analysis of CGE against the best results from three baselines across three GNN backbones and four datasets. This analysis aims to verify CGE's model utility in comparison to \textit{Scratch}, as referenced in the Preliminaries. 

\paragraph{Results.} 
The experimental results presented in Table \ref{tab:metrics} reveal the following findings: (1) CGE outperforms all established baselines in terms of model utility. For instance, on Reddit, CGE shows an average increase of 11.56\% compared to GUIDE and an even more significant average increase of 33.24\% compared to GraphEraser. This superiority is attributed to two factors:
(a) CGE employs a graph-optimized community detection method, resulting in a more accurate community distribution than the balanced partitioning used in the BP-SM-TA paradigm.
(b) CGE accounts for the inherent relationships between subgraphs, reducing information loss, a consideration overlooked by the BP-SM-TA paradigm. (2) CGE demonstrates comparable model utility to \textit{Scratch}, which serves as the foundational baseline due to its broad and outstanding performance achieved through the entire retraining of the dataset. Notably, CGE has even outperformed \textit{Scratch} in certain cases, primarily due to its pre-integration of similar data within the same community in the original graph, effectively reducing noise. This results in a newly mapped graph with more robust and accurate information. CGE achieves average performance closer to \textit{Scratch}, fulfilling the objectives of unlearning.

\subsection{Efficiency of CGE}
To answer the question (2), we evaluate CGE's efficiency in two tasks: model deployment and unlearning.

In the \textit{\textbf{model deployment stage}}, the unlearning framework generates a complete prediction model, including graph partitioning, data pre-processing, and initial model training. In the \textit{\textbf{unlearning stage}}, the model performs deterministic unlearning on a demand sequence and produces an updated model based on the deployed one. Since the training time across three backbones on the same dataset is linear, we show the average efficiency of different unlearning frameworks across all backbones, as presented in Table  \ref{tab:eff}.

During \textit{\textbf{model deployment stage}}, CGE's efficiency surpasses most baselines by retaining only the most representative information in the mapped graph. Compared to unlearning methods trained on the original graph, the mapped graph offers smaller-scale training data with higher representativeness. For example, as shown in Table \ref{tab:com}, the CS dataset requires fewer training data and parameters for the mapped graph compared to the representative method GraphEraser under the BP-SM-TA paradigm.

During \textit{\textbf{unlearning stage}}, CGE's efficiency advantage becomes even more pronounced, significantly surpassing Scratch and all baselines. The graph structure mapping minimizes the impact of data size differences, allowing CGE to reduce unlearning time to seconds without sacrificing model utility. This efficiency results from both the exponential reduction in data and parameters and the node-level unlearning strategy. Unlike traditional methods that require retraining and aggregation of submodels, CGE only needs to recalculate the relevant mapped node-level data, significantly reducing time complexity.

\begin{table}[htbp]\small
\centering
\newcolumntype{Z}{>{\centering\let\newline\\\arraybackslash\hspace{0pt}}X}
\begin{tabularx}{\columnwidth}{l|ZZZ}
    \toprule[1.5pt]
    \textbf{Index} & \textbf{\text{GraphEraser}} & \textbf{CGE} \\
    \midrule
    \textbf{\#Nodes (T)} & 18,333 & 1,756\\
    \textbf{\#Parameters (T)} & 43,498,280 & 217,875\\
    \textbf{\#Parameters (U)} & 4,890,365 & 217,875\\
    \bottomrule[1.5pt]
\end{tabularx}
\caption{Comparison of the quantity of training nodes and parameters across different unlearning frameworks on CS Dataset. Where `T' and `U' denote the training and unlearning processes, respectively.}
\label{tab:com}
\end{table}

\subsection{Unlearning Ability}
To determine the effectiveness of CGE in unlearning, we utilize a node-level GNN member inference attack methodology (MIA) \cite{mia}. This approach involves calculating the discrepancy between the original model and the model after CGE unlearning, to determine whether a node has been successfully removed from the model. After randomly removing 0.5\% of all nodes according to the experimental parameters, the MIA attack was executed. We consider two scenarios: $\mathcal{A}$ and $\mathcal{A_C}$, representing the MIA attack results on the original GNN backbones and CGE, respectively. As shown in Table \ref{tab:mia}, the AUC of MIA on CGE is significantly lower than that of the original GNN backbones, approaching 0.5, which is analogous to random guessing. This indicates that CGE does not produce additional information leakage and that its unlearning of nodes and associated effects is effective. 

\begin{table}[htbp]
\centering

\small
\newcolumntype{Z}{>{\centering\let\newline\\\arraybackslash\hspace{0pt}}X}
\begin{tabularx}{\columnwidth}{l|ZZZZZZ}
    \toprule[1.5pt]
    \multirow{2}{*}{\textbf{Dataset}} & \multicolumn{2}{c}{\textbf{GCN}} & \multicolumn{2}{c}{\textbf{GAT}} & \multicolumn{2}{c}{\textbf{SAGE}}\\
    & $\mathcal{A}$ & $\mathcal{A_C}$ & $\mathcal{A}$&$\mathcal{A_C}$&$\mathcal{A}$&$\mathcal{A_C}$ \\
    \midrule
    \textbf{Cora} & \text{73.3} & \text{50.1} & \text{75.2} & \text{49.8} & \text{74.1} & \text{52.1}  \\
    \textbf{Citeseer} & \text{78.3} & \text{52.1} & \text{79.1} & \text{50.2} & \text{79.7} & \text{50.7}  \\
    \textbf{CS} & \text{73.1} & \text{50.1} & \text{72.3} & \text{49.3} & \text{70.8} & \text{48.8}  \\
    \textbf{Reddit} & \text{63.1} & \text{50.0} & \text{65.5} & \text{49.6} & \text{65.1} & \text{49.8}  \\
    \bottomrule[1.5pt]
\end{tabularx}
\caption{Attack AUC of membership inference against GNN backbones and CGE (\%).}
\label{tab:mia}
\end{table}

\subsection{Ablation Study}
We aim to evaluate the contribution of different modules in CGE to the system by addressing the following questions:

\paragraph{Compared to previous schemes, is overlapping community detection more advantageous?} Most previous schemes use similar node feature clustering methods for subgraph-balanced partitioning, as BP-SM-TA requires each subgraph to represent the original graph. In contrast, CGE calculates a highly representative mapped graph without cutting the original graph, thereby limiting information loss. To evaluate the schemes for CGE, we use a custom metric, \textit{Information Retention}, which measures cosine similarity between the low-dimensional embeddings of the original graph and the subgraph—higher values indicate less information loss. Additional details and validation experiments are provided in Appendix C.  Figure \ref{fig:a} shows the evaluation results of different graph partitioning schemes Balanced Kmeans (BEKM) and Louvain-OSLOM (OSLOM) across four datasets. The balanced partitioning scheme is suboptimal, especially in the Reddit dataset, whose original graph representativeness is below zero, explaining why GraphEraser performs poorly on Reddit. In contrast, overlapping community detection better suits CGE. Figure \ref{fig:b} shows that BEKM leads to a significant decrease in CGE's performance, further validating the above explanation.

\begin{figure}[t]
    \centering
    \subfigure[]{
    \includegraphics[width=0.47\columnwidth]{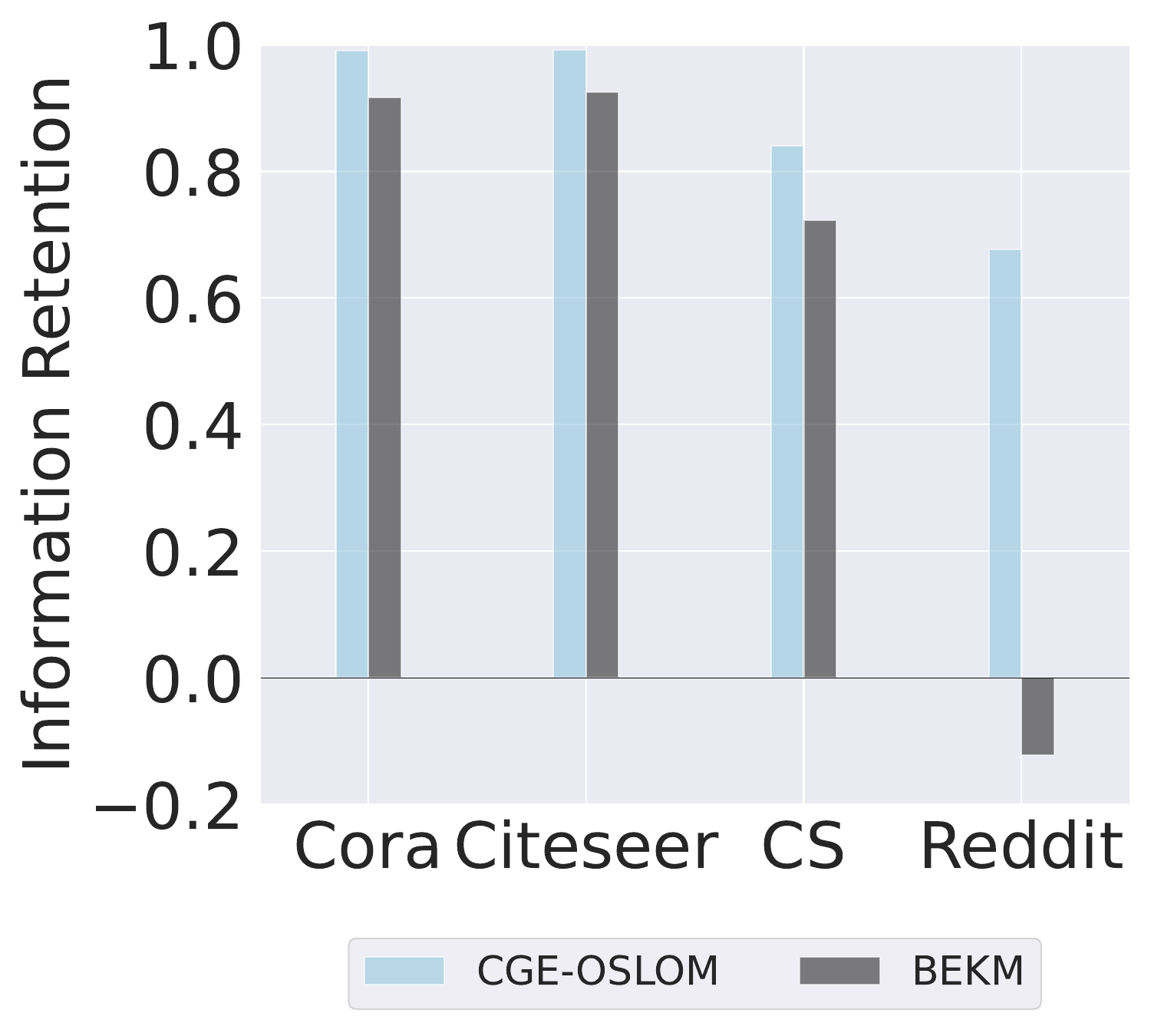}
    \label{fig:a}
    }
    \subfigure[]{
    \includegraphics[width=0.47\columnwidth]{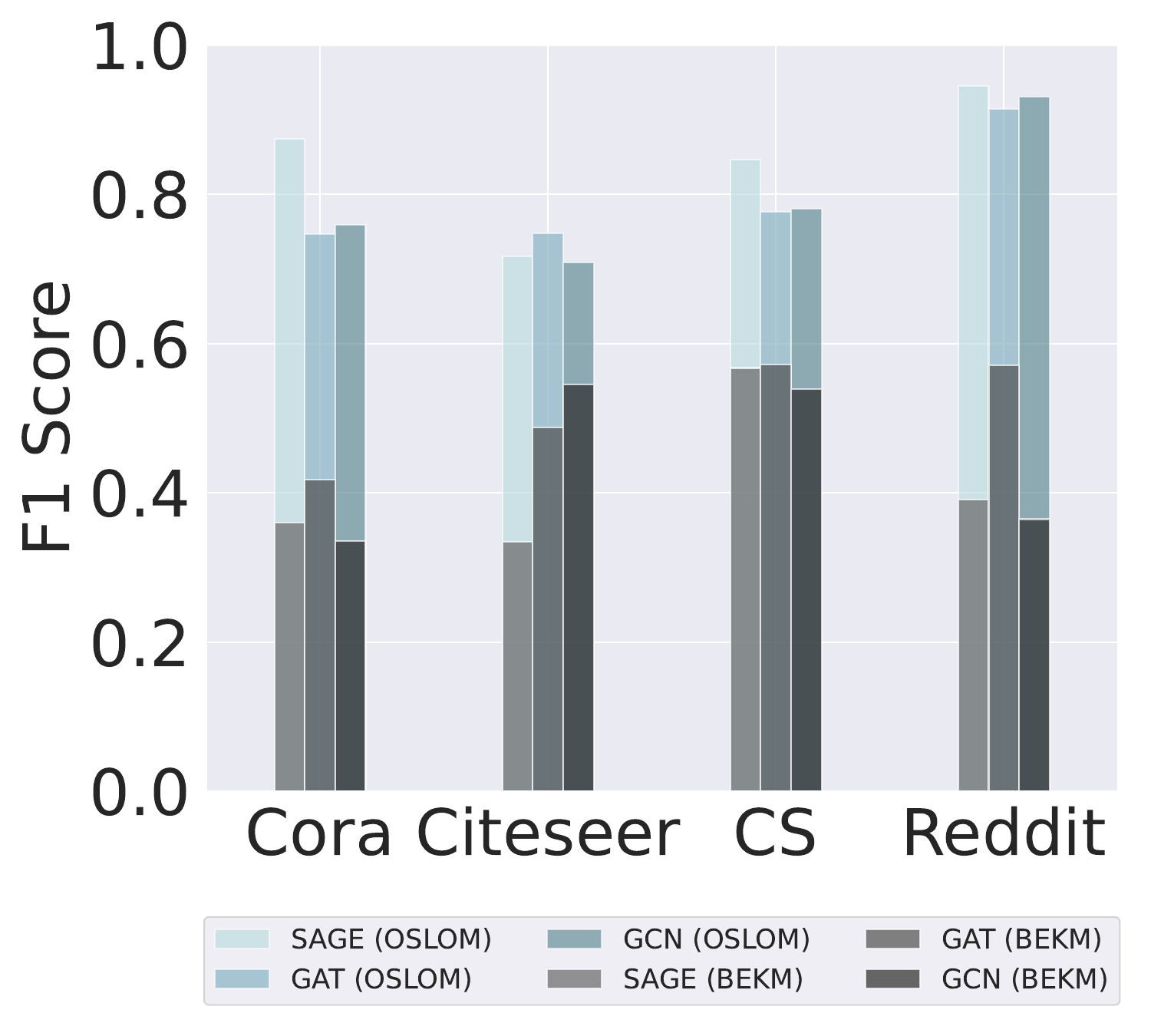}
    \label{fig:b}
    }
    \caption{Quality and utility evaluation of different graph partitioning schemes.}
    \label{fig:enter-label}
\end{figure}

\begin{figure}[t]
\centering
\includegraphics[width=0.98\columnwidth]{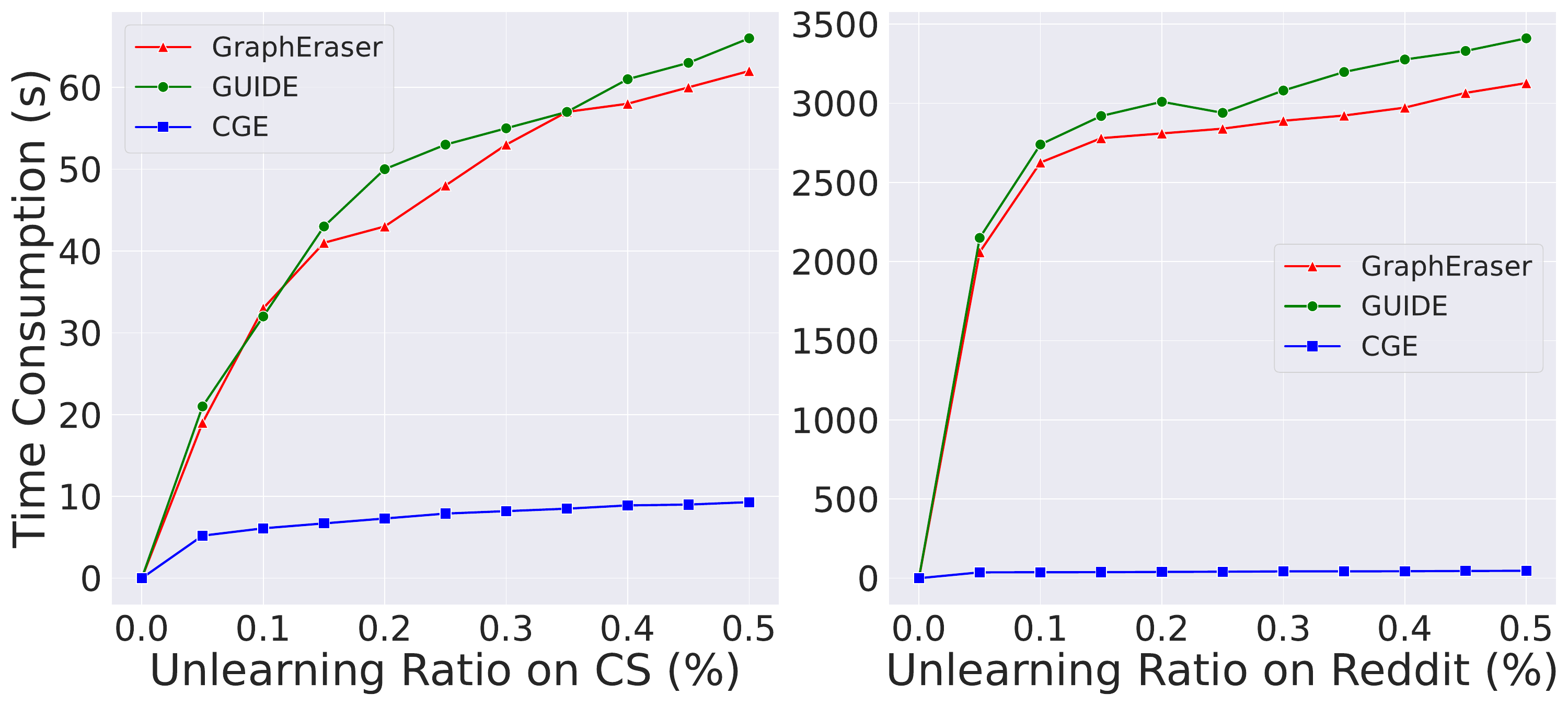} 
\caption{Time consumption of each method under different unlearning node ratios.}
\label{fig:ratio_com}
\end{figure}

\paragraph{Why can CGE overcome the limitations of balanced partitioning?} Previous experiments have shown that balanced partitioning is a suboptimal graph partitioning scheme, as it causes significant information loss due to graph cutting. This scheme is adopted by the BP-SM-TA paradigm to prevent imbalanced submodels from causing unmanageable learning time consumption, but it sacrifices model utility in the process. In contrast, CGE, based on the GSMU paradigm, offers a node-level unlearning solution with low parameter count and complexity. Figure \ref{fig:ratio_com} shows the unlearning efficiency of CGE on two large-scale datasets CS and Reddit under different unlearning node ratios. Notably, CGE avoids exponential increases in unlearning time as the number of forgotten nodes grows, thus eliminating the need for balanced partitioning as a protective measure.

\section{Related Work}
Machine unlearning aims to erase specific data and its derived effects. Following Cao's work~\cite{caotowards}, this concept has evolved toward the development of both approximate and deterministic unlearning methods.

Approximate unlearning~\cite{gif, approxi2} achieves statistical unlearning by fine-tuning model parameters, but it cannot guarantee precise data erasure, posing challenges for non-convex models like deep learning. Conversely, deterministic unlearning ensures the complete removal of data points and their influence, often through retraining. SISA~\cite{sisa} proposed a method involving random balanced partitioning into multiple sub-models, where only the affected sub-models are retrained based on unlearning requests, and the final prediction is aggregated from these sub-models. This approach is referred to as the BP-SM-TA paradigm in this paper. Building on this, Chen et al. extended the concept to graph-structured data with GraphEraser~\cite{grapheraser}, and the recently proposed GUIDE~\cite{guide} further optimized it for inductive graphs.

In contrast to previous work, the proposed CGE framework employs a novel graph structure mapping paradigm, which maps the original graph to a highly representative low-order graph. It introduces a low-parameter node-level unlearning strategy to enhance the utilization and efficiency of the graph structure within the unlearning framework.

\section{Conclusions}

In this work, we introduce GSMU, a novel graph structure mapping unlearning paradigm that offers an intuitive and efficient approach to graph unlearning. We propose the Community-Centric Graph Eraser (CGE) as a practical implementation of GSMU. CGE first detects communities and maps them to nodes, creating a mapped graph. This graph is then used to reconstruct node-level unlearning operations and representation learning tasks, resulting in high graph structure utilization and significantly reduced unlearning computational costs. Comprehensive evaluations demonstrate that CGE effectively leverages graph structures, offering excellent model utility and significantly improved unlearning efficiency. This work establishes a groundbreaking paradigm for graph unlearning with its streamlined architecture and superior performance.

CGE shows strong potential but faces challenges with traditional community detection techniques on large-scale graphs. Future work will leverage deep learning-based methods to enhance data insights while addressing privacy concerns, aiming for a balanced and effective approach.

\section{Acknowledgments}
This work was partially supported by the Project of Guangxi Science and Technology (GuiKeAB23026040), Research Fund of Guangxi Key Lab of Multi-source Information Mining \& Security (MIMS24-13), Research Fund of Guangxi Key Lab of Multi-source Information Mining \& Security (24-A-01-02), and Innovation Project of Guangxi Graduate Education (XYCSR2024102). 

\bibliography{aaai25}

\begin{thebibliography}{37}
\providecommand{\natexlab}[1]{#1}

\bibitem[{Blondel et~al.(2008)Blondel, Guillaume, Lambiotte, and Lefebvre}]{louvain}
Blondel, V.~D.; Guillaume, J.-L.; Lambiotte, R.; and Lefebvre, E. 2008.
\newblock Fast unfolding of communities in large networks.
\newblock \emph{Journal of statistical mechanics: theory and experiment}, 2008(10): P10008.

\bibitem[{Bourtoule et~al.(2021)Bourtoule, Chandrasekaran, Choquette-Choo, Jia, Travers, Zhang, Lie, and Papernot}]{sisa}
Bourtoule, L.; Chandrasekaran, V.; Choquette-Choo, C.~A.; Jia, H.; Travers, A.; Zhang, B.; Lie, D.; and Papernot, N. 2021.
\newblock Machine unlearning.
\newblock In \emph{2021 IEEE Symposium on Security and Privacy (SP)}, 141--159. IEEE.

\bibitem[{Cao and Yang(2015)}]{caotowards}
Cao, Y.; and Yang, J. 2015.
\newblock Towards making systems forget with machine unlearning.
\newblock In \emph{2015 IEEE symposium on security and privacy}, 463--480. IEEE.

\bibitem[{Chen et~al.(2022{\natexlab{a}})Chen, Sun, Zhang, and Ding}]{receraser}
Chen, C.; Sun, F.; Zhang, M.; and Ding, B. 2022{\natexlab{a}}.
\newblock Recommendation unlearning.
\newblock In \emph{Proceedings of the ACM Web Conference 2022}, 2768--2777.

\bibitem[{Chen et~al.(2022{\natexlab{b}})Chen, Zhang, Wang, Backes, Humbert, and Zhang}]{grapheraser}
Chen, M.; Zhang, Z.; Wang, T.; Backes, M.; Humbert, M.; and Zhang, Y. 2022{\natexlab{b}}.
\newblock Graph unlearning.
\newblock In \emph{Proceedings of the 2022 ACM SIGSAC conference on computer and communications security}, 499--513.

\bibitem[{Cofone(2020)}]{law3}
Cofone, I. 2020.
\newblock \emph{The right to be forgotten: A Canadian and comparative perspective}.
\newblock Routledge.

\bibitem[{Fan et~al.(2019)Fan, Ma, Li, He, Zhao, Tang, and Yin}]{rec}
Fan, W.; Ma, Y.; Li, Q.; He, Y.; Zhao, E.; Tang, J.; and Yin, D. 2019.
\newblock Graph neural networks for social recommendation.
\newblock In \emph{The world wide web conference}, 417--426.

\bibitem[{Guo et~al.(2019)Guo, Goldstein, Hannun, and Van Der~Maaten}]{approxi2}
Guo, C.; Goldstein, T.; Hannun, A.; and Van Der~Maaten, L. 2019.
\newblock Certified data removal from machine learning models.
\newblock \emph{arXiv preprint arXiv:1911.03030}.

\bibitem[{Hamilton, Ying, and Leskovec(2017)}]{sage}
Hamilton, W.; Ying, Z.; and Leskovec, J. 2017.
\newblock Inductive representation learning on large graphs.
\newblock \emph{Advances in neural information processing systems}, 30.

\bibitem[{He et~al.(2021)He, Wen, Wu, Backes, Shen, and Zhang}]{mia}
He, X.; Wen, R.; Wu, Y.; Backes, M.; Shen, Y.; and Zhang, Y. 2021.
\newblock Node-level membership inference attacks against graph neural networks.
\newblock \emph{arXiv preprint arXiv:2102.05429}.

\bibitem[{Izzo et~al.(2021)Izzo, Smart, Chaudhuri, and Zou}]{approxi3}
Izzo, Z.; Smart, M.~A.; Chaudhuri, K.; and Zou, J. 2021.
\newblock Approximate data deletion from machine learning models.
\newblock In \emph{International Conference on Artificial Intelligence and Statistics}, 2008--2016. PMLR.

\bibitem[{Jin et~al.(2021)Jin, Yu, Jiao, Pan, He, Wu, Philip, and Zhang}]{cdsurvey1}
Jin, D.; Yu, Z.; Jiao, P.; Pan, S.; He, D.; Wu, J.; Philip, S.~Y.; and Zhang, W. 2021.
\newblock A survey of community detection approaches: From statistical modeling to deep learning.
\newblock \emph{IEEE Transactions on Knowledge and Data Engineering}, 35(2): 1149--1170.

\bibitem[{Khan and Niazi(2017)}]{cdsurvey2}
Khan, B.~S.; and Niazi, M.~A. 2017.
\newblock Network community detection: A review and visual survey.
\newblock \emph{arXiv preprint arXiv:1708.00977}.

\bibitem[{Kipf and Welling(2016{\natexlab{a}})}]{gcn}
Kipf, T.~N.; and Welling, M. 2016{\natexlab{a}}.
\newblock Semi-supervised classification with graph convolutional networks.
\newblock \emph{arXiv preprint arXiv:1609.02907}.

\bibitem[{Kipf and Welling(2016{\natexlab{b}})}]{vgae}
Kipf, T.~N.; and Welling, M. 2016{\natexlab{b}}.
\newblock Variational graph auto-encoders.
\newblock \emph{arXiv preprint arXiv:1611.07308}.

\bibitem[{Lancichinetti et~al.(2011)Lancichinetti, Radicchi, Ramasco, and Fortunato}]{oslom}
Lancichinetti, A.; Radicchi, F.; Ramasco, J.~J.; and Fortunato, S. 2011.
\newblock Finding statistically significant communities in networks.
\newblock \emph{PloS one}, 6(4): e18961.

\bibitem[{Li et~al.(2024{\natexlab{a}})Li, Cheng, Zhang, and Zhang}]{lic}
Li, C.; Cheng, D.; Zhang, G.; and Zhang, S. 2024{\natexlab{a}}.
\newblock Contrastive learning for fair graph representations via counterfactual graph augmentation.
\newblock \emph{Knowledge-Based Systems}, 305: 112635.

\bibitem[{Li et~al.(2024{\natexlab{b}})Li, Chen, Xi, Huang, Zhang, and Zhang}]{liz}
Li, Z.; Chen, B.; Xi, P.; Huang, Z.; Zhang, W.; and Zhang, S. 2024{\natexlab{b}}.
\newblock Spatio-Temporal Dynamically Fused Graph Convolutional Network.
\newblock In \emph{2024 International Joint Conference on Neural Networks (IJCNN)}, 1--8. IEEE.

\bibitem[{Li et~al.(2022)Li, Jian, Wang, and Chen}]{ccgnn}
Li, Z.; Jian, X.; Wang, Y.; and Chen, L. 2022.
\newblock Cc-gnn: A community and contraction-based graph neural network.
\newblock In \emph{2022 IEEE International Conference on Data Mining (ICDM)}, 231--240. IEEE.

\bibitem[{Mantelero(2013)}]{law2}
Mantelero, A. 2013.
\newblock The EU Proposal for a General Data Protection Regulation and the roots of the ‘right to be forgotten’.
\newblock \emph{Computer Law \& Security Review}, 29(3): 229--235.

\bibitem[{Nguyen et~al.(2022)Nguyen, Huynh, Nguyen, Liew, Yin, and Nguyen}]{musurvey}
Nguyen, T.~T.; Huynh, T.~T.; Nguyen, P.~L.; Liew, A. W.-C.; Yin, H.; and Nguyen, Q. V.~H. 2022.
\newblock A survey of machine unlearning.
\newblock \emph{arXiv preprint arXiv:2209.02299}.

\bibitem[{Pardau(2018)}]{law1}
Pardau, S.~L. 2018.
\newblock The california consumer privacy act: Towards a european-style privacy regime in the united states.
\newblock \emph{J. Tech. L. \& Pol'y}, 23: 68.

\bibitem[{Qiu et~al.(2018)Qiu, Tang, Ma, Dong, Wang, and Tang}]{smedia}
Qiu, J.; Tang, J.; Ma, H.; Dong, Y.; Wang, K.; and Tang, J. 2018.
\newblock Deepinf: Social influence prediction with deep learning.
\newblock In \emph{Proceedings of the 24th ACM SIGKDD international conference on knowledge discovery \& data mining}, 2110--2119.

\bibitem[{Rosvall and Bergstrom(2008)}]{infomap}
Rosvall, M.; and Bergstrom, C.~T. 2008.
\newblock Maps of random walks on complex networks reveal community structure.
\newblock \emph{Proceedings of the national academy of sciences}, 105(4): 1118--1123.

\bibitem[{Said et~al.(2023)Said, Derr, Shabbir, Abbas, and Koutsoukos}]{gusurvey}
Said, A.; Derr, T.; Shabbir, M.; Abbas, W.; and Koutsoukos, X. 2023.
\newblock A survey of graph unlearning.
\newblock \emph{arXiv preprint arXiv:2310.02164}.

\bibitem[{Shchur et~al.(2018{\natexlab{a}})Shchur, Mumme, Bojchevski, and G{\"u}nnemann}]{cs}
Shchur, O.; Mumme, M.; Bojchevski, A.; and G{\"u}nnemann, S. 2018{\natexlab{a}}.
\newblock Pitfalls of graph neural network evaluation.
\newblock \emph{arXiv preprint arXiv:1811.05868}.

\bibitem[{Shchur et~al.(2018{\natexlab{b}})Shchur, Mumme, Bojchevski, and G{\"u}nnemann}]{f1}
Shchur, O.; Mumme, M.; Bojchevski, A.; and G{\"u}nnemann, S. 2018{\natexlab{b}}.
\newblock Pitfalls of graph neural network evaluation.
\newblock \emph{arXiv preprint arXiv:1811.05868}.

\bibitem[{Veli{\v{c}}kovi{\'c} et~al.(2017)Veli{\v{c}}kovi{\'c}, Cucurull, Casanova, Romero, Lio, and Bengio}]{gat}
Veli{\v{c}}kovi{\'c}, P.; Cucurull, G.; Casanova, A.; Romero, A.; Lio, P.; and Bengio, Y. 2017.
\newblock Graph attention networks.
\newblock \emph{arXiv preprint arXiv:1710.10903}.

\bibitem[{Wang, Huai, and Wang(2023)}]{guide}
Wang, C.-L.; Huai, M.; and Wang, D. 2023.
\newblock Inductive graph unlearning.
\newblock In \emph{32nd USENIX Security Symposium (USENIX Security 23)}, 3205--3222.

\bibitem[{Wei et~al.(2024)Wei, Yuan, Fu, Sun, Peng, Li, and Hu}]{fux}
Wei, Y.; Yuan, H.; Fu, X.; Sun, Q.; Peng, H.; Li, X.; and Hu, C. 2024.
\newblock Poincar{\'e} Differential Privacy for Hierarchy-aware Graph Embedding.
\newblock In \emph{Proceedings of the AAAI Conference on Artificial Intelligence}, volume~38, 9160--9168.

\bibitem[{Wu et~al.(2023)Wu, Yang, Qian, Sui, Wang, and He}]{gif}
Wu, J.; Yang, Y.; Qian, Y.; Sui, Y.; Wang, X.; and He, X. 2023.
\newblock Gif: A general graph unlearning strategy via influence function.
\newblock In \emph{Proceedings of the ACM Web Conference 2023}, 651--661.

\bibitem[{Xie, Szymanski, and Liu(2011)}]{slpa}
Xie, J.; Szymanski, B.~K.; and Liu, X. 2011.
\newblock Slpa: Uncovering overlapping communities in social networks via a speaker-listener interaction dynamic process.
\newblock In \emph{2011 ieee 11th international conference on data mining workshops}, 344--349. IEEE.

\bibitem[{Xu et~al.(2018)Xu, Hu, Leskovec, and Jegelka}]{gin}
Xu, K.; Hu, W.; Leskovec, J.; and Jegelka, S. 2018.
\newblock How powerful are graph neural networks?
\newblock \emph{arXiv preprint arXiv:1810.00826}.

\bibitem[{Yan et~al.(2022)Yan, Li, Guo, Li, Li, and Lin}]{arcane}
Yan, H.; Li, X.; Guo, Z.; Li, H.; Li, F.; and Lin, X. 2022.
\newblock ARCANE: An Efficient Architecture for Exact Machine Unlearning.
\newblock In \emph{IJCAI}, volume~6, 19.

\bibitem[{Yang, Cohen, and Salakhudinov(2016)}]{cc}
Yang, Z.; Cohen, W.; and Salakhudinov, R. 2016.
\newblock Revisiting semi-supervised learning with graph embeddings.
\newblock In \emph{International conference on machine learning}, 40--48. PMLR.

\bibitem[{Zhang et~al.(2025)Zhang, Yuan, Cheng, Liu, Li, and Zhang}]{guix}
Zhang, G.; Yuan, G.; Cheng, D.; Liu, L.; Li, J.; and Zhang, S. 2025.
\newblock Disentangled contrastive learning for fair graph representations.
\newblock \emph{Neural Networks}, 181: 106781.

\bibitem[{Zhang et~al.(2017)Zhang, Li, Zong, Zhu, and Cheng}]{zhangknn}
Zhang, S.; Li, X.; Zong, M.; Zhu, X.; and Cheng, D. 2017.
\newblock Learning k for knn classification.
\newblock \emph{ACM Transactions on Intelligent Systems and Technology (TIST)}, 8(3): 1--19.

\end{thebibliography}

\clearpage
\appendix

\section{A. GNN Backbones}
The experimental verification of this work is based on three GNN backbones, which are representative mainstream graph neural network models, to verify the applicability of CGE.

\paragraph{Graph Convolutional Network (GCN):} GCN~\cite{gcn} uses spectral graph convolutions to aggregate features from neighbors. The update rule is:
    \begin{equation}
        \mathbf{H}^{(l+1)} = \sigma\left(\hat{\mathbf{D}}^{-\frac{1}{2}} \hat{\mathbf{A}} \hat{\mathbf{D}}^{-\frac{1}{2}} \mathbf{H}^{(l)} \mathbf{W}^{(l)}\right),
    \end{equation}
where \( \hat{\mathbf{A}} = \mathbf{A} + \mathbf{I} \) and \( \hat{\mathbf{D}} \) is its degree matrix.

\paragraph{Graph Attention Network (GAT):} GAT~\cite{gat} introduces attention by weighting neighbor features based on their importance. The attention coefficient is:
   \begin{equation}
       \alpha_{ij} = \frac{\exp\left(\text{LeakyReLU}\left(\mathbf{a}^\top [\mathbf{W}\mathbf{h}_i \| \mathbf{W}\mathbf{h}_j]\right)\right)}{\sum_{k \in \mathcal{N}(i)} \exp\left(\text{LeakyReLU}\left(\mathbf{a}^\top [\mathbf{W}\mathbf{h}_i \| \mathbf{W}\mathbf{h}_k]\right)\right)}.
   \end{equation}

\paragraph{Graph SAmple and aggreGatE (SAGE):} SAGE~\cite{sage} generates node embeddings by sampling and aggregating neighbor features. The update rule is:
\begin{equation}
    \mathbf{h}_v^{(l+1)} = \sigma\left(\mathbf{W}^{(l)} \cdot \text{AGG}\left(\{\mathbf{h}_v^{(l)}\} \cup \{\mathbf{h}_u^{(l)}, \forall u \in \mathcal{N}(v)\}\right)\right),
\end{equation}
where \( \text{AGG} \) can be mean, LSTM, or pooling. 

\section{B. Specific Experimental Settings}
\subsection{Parameter Settings}
Since the three baselines are based on the BP-SM-TA paradigm, we set the same number of shards for each. We used the Adam optimizer with a default learning rate of 0.01 and a weight decay of 0.001. As a result, Cora, Citeseer, CS, and Reddit were divided into 20, 20, 80, and 300 shards, respectively. Each GNN backbone was trained for 200 epochs. We ran all experiments related to model utility and efficiency 10 times and reported the average results.

\subsection{Evaluation Metrics}
We evaluate the graph model's utility with the Macro F1 score and assess the unlearning capability using node-level GNN membership inference attacks, measured by the Area Under the Curve (AUC).
\paragraph{Macro F1 Score.} The Macro F1 score averages the F1 scores for each class, providing an overall measure of the model's classification performance across all classes~\cite{f1}. It is particularly useful when dealing with imbalanced data, ensuring that each class is equally represented in the final score.
\paragraph{AUC (Area Under the Curve).}AUC evaluates the effectiveness of the unlearning process by measuring the model's susceptibility to membership inference attacks. It quantifies how well the model can distinguish between nodes that were part of the training set and those that were not, with higher AUC values indicating weaker unlearning ability.

\section{C. Further Ablation Study}
The BP-SM-TA paradigm requires that each balanced subgraph must maintain a significant degree of representativeness of the original graph. We believe this enforced constraint severely compromises the performance of the graph structure. To validate our hypothesis, we evaluate the subgraph (community) partitioning results from two perspectives: \textbf{\textit{Information Retention}} and \textbf{\textit{Conductance}}. The evaluation of Information Retention has been presented in the main text, and we will briefly explain its underlying principles here.
\subsection{Information Retention}
In the ablation study, we assess the Information Retention across different graph partitioning schemes to measure how well the subgraphs represent the original graph. Specifically, We compute the embeddings $\mathbf{Z}_{\text{orig}}$ and $\mathbf{Z}_{\text{sub}}$ for the original graph and the subgraph, respectively, using a Graph Autoencoder~\cite{vgae}:
\begin{equation}
    \mathbf{Z}_{\text{orig}} = \text{Encoder}(\mathbf{X}_{\text{orig}}),
\end{equation}
\begin{equation}
    \mathbf{Z}_{\text{sub}} = \text{Encoder}(\mathbf{X}_{\text{sub}}).
\end{equation}
The autoencoder reconstructs the features from these embeddings:
\begin{equation}
    \hat{\mathbf{X}}_{\text{orig}} = \text{Decoder}(\mathbf{Z}_{\text{orig}}),
\end{equation}
\begin{equation}
    \hat{\mathbf{X}}_{\text{sub}} = \text{Decoder}(\mathbf{Z}_{\text{sub}}).
\end{equation}
Information retention is measured by the cosine similarity between the reconstructed features of the original graph and the mean of the subgraph reconstruction:
\begin{equation}
    \text{InfoRet} = \frac{\hat{\mathbf{X}}_{\text{orig}} \cdot \left(\frac{1}{|\mathcal{V}_{\text{sub}}|} \sum_{v \in \mathcal{V}_{\text{sub}}} \hat{\mathbf{X}}_{\text{sub}, v}\right)}{\|\hat{\mathbf{X}}_{\text{orig}}\| \left\|\frac{1}{|\mathcal{V}_{\text{sub}}|} \sum_{v \in \mathcal{V}_{\text{sub}}} \hat{\mathbf{X}}_{\text{sub}, v}\right\|}
\end{equation}
where \(\mathcal{V}_{\text{sub}}\) represents the set of nodes in the subgraph.

The final information retention score is the mean similarity:
\begin{equation}
    \text{InfoRet\_score} = \frac{1}{N} \sum_{i=1}^{N} \text{InfoRet}_i,
\end{equation}
where $N$ is the number of nodes or comparisons. The value of information retention is defined between $-1$ and $1$, with a higher value indicating that the information characteristics of the subgraph are more similar to those of the original graph, and the information loss is lower.

\subsection{Conductance}
Conductance is a key metric used to evaluate the quality of community structures in a graph. It measures the ratio of the number of edges that cut through the boundary of a community to the total number of edges within that community. Lower conductance values indicate that the community is more tightly knit, with fewer edges crossing the boundary, thus suggesting a more compact and well-defined community structure.

The conductance $\phi(C)$ of a community $C$ is given by:

\begin{equation}
    \phi(C) = \frac{\text{cut}(C, \bar{C})}{\min(\text{vol}(C), \text{vol}(\bar{C}))},
\end{equation}
where:
\begin{itemize}
    \item $\text{cut}(C, \bar{C})$ is the number of edges connecting nodes in $S$ to nodes outside of $C$,
    \item $\text{vol}(C)$ is the sum of the degrees of nodes within $C$,
    \item $\text{vol}(\bar{C})$ is the volume of the complement of $C$.
\end{itemize}

Lower values of $\phi(C)$ indicate a more cohesive community with fewer external connections relative to its internal structure.

\begin{figure}
    \centering
    \includegraphics[width=0.9\columnwidth]{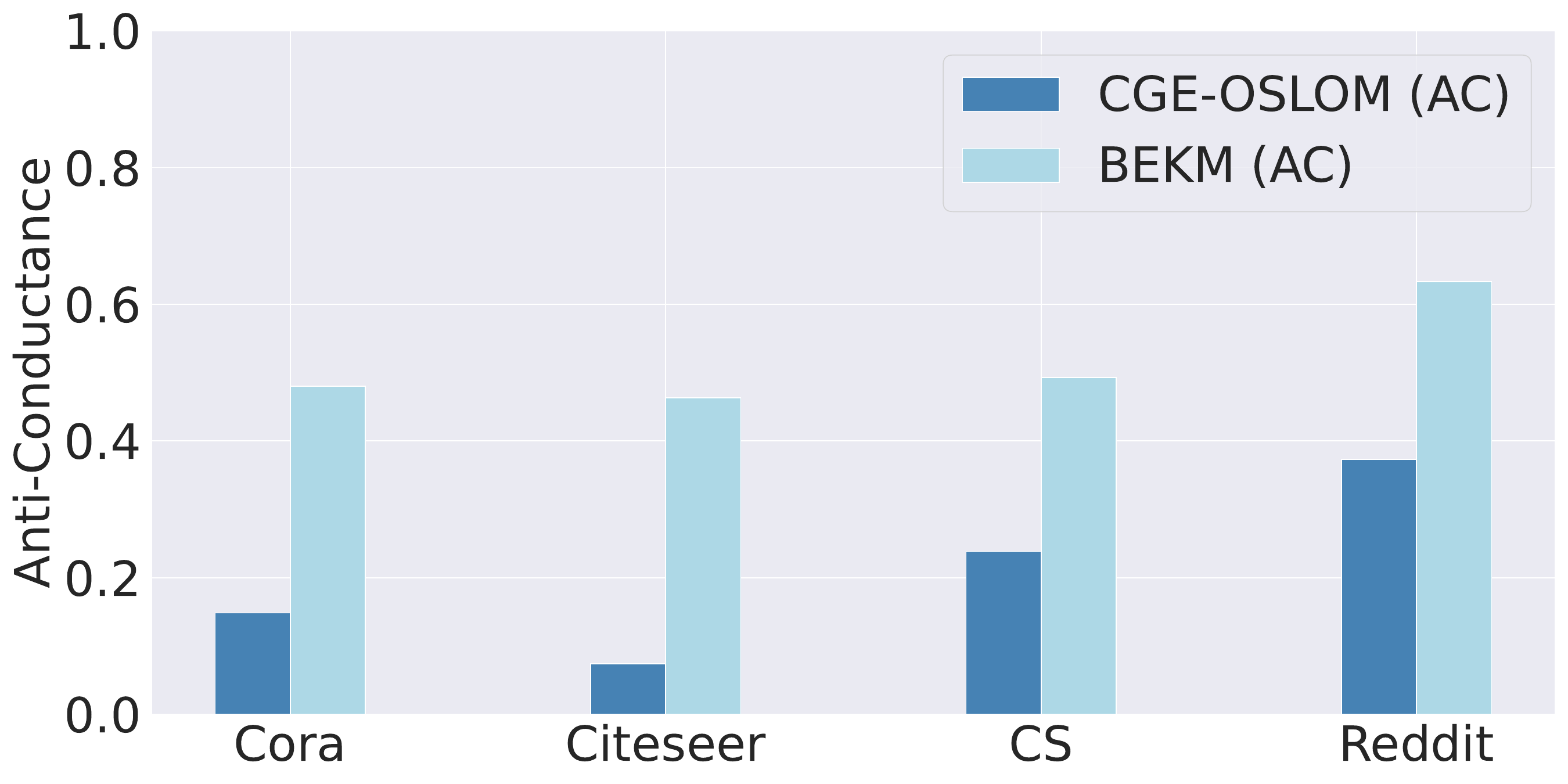}
    \caption{Conductance evaluation of different graph partitioning schemes.}
    \label{fig:con}
\end{figure}

The conductance evaluation of the balanced community partitioning scheme BEKM and the Louvain-OSLOM (referred to as OSLOM) method used in this work is shown in Figure \ref{fig:con}. OSLOM achieves a lower conductance score, demonstrating that the communities it partitions are more compact internally and sparser externally. This indicates that OSLOM better preserves the intrinsic structural features of communities while reducing external connections, thereby improving community purity and structural tightness. In contrast, BEKM shows higher conductance scores, indicating more boundary connections and looser internal structures in its partitioned communities. This difference highlights the superiority of OSLOM for complex network analysis, making it a more suitable approach for the CGE graph unlearning tasks based on GSMU. 

\section{D. Adaptability of Cmmunity Detection Methods}
We introduce two widely used community detection methods: Infomap~\cite{infomap} and SLPA~\cite{slpa}.

\begin{itemize}
    \item \textbf{Infomap.} Infomap utilizes a method based on random walks and information theory to detect communities. It minimizes the description length of a network by finding an optimal partition that compresses the network’s structure. The objective function for Infomap can be expressed as:
\begin{equation}
    \text{Modularity} = \frac{1}{2m} \sum_{i,j} \left( A_{ij} - \frac{k_i k_j}{2m} \right) \delta(c_i, c_j),
\end{equation}
where \(A_{ij}\) is the adjacency matrix, \(k_i\) and \(k_j\) are the degrees of nodes \(i\) and \(j\), \(m\) is the total number of edges, and \(\delta(c_i, c_j)\) is 1 if nodes \(i\) and \(j\) are in the same community and 0 otherwise.
    \item \textbf{SLPA.} SLPA (Speaker-Listener Label Propagation Algorithm) is another effective method that uses a label propagation approach to identify communities. Each node in the network iteratively updates its label based on the labels of its neighbors. The update rule for SLPA is given by:
\begin{equation}
    l_i(t+1) = \text{mode}\{ l_j(t) \mid j \in N(i) \},
\end{equation}
where \(l_i(t)\) is the label of node \(i\) at iteration \(t\), \(N(i)\) denotes the set of neighbors of node \(i\), and \(\text{mode}\) denotes the most frequent label among the neighbors.
\end{itemize}

\begin{table}[htbp]\footnotesize
\centering

\newcolumntype{Z}{>{\centering\let\newline\\\arraybackslash\hspace{0pt}}X}
\begin{tabularx}{\columnwidth}{lZZZ}
    \toprule[1.5pt]
    \multirow{2}{*}{\textbf{Dataset}} & \multicolumn{3}{c}{\textbf{Method}} \\
    \cmidrule(lr){2-4}
    & \textbf{SLPA} & \textbf{OSLOM} & \textbf{Infomap} \\
    \midrule
    \textbf{Cora} & 16 & 83 & 5 \\
    \textbf{Citeseer} & 18 & 99 & 6 \\
    \textbf{CS} & 172 & 747 & 77\\
    \textbf{Reddit} & - & 18,077 & 8,763\\
        
    \bottomrule[1.5pt]
\end{tabularx}
\caption{Partition time consumption of different community detection methods. (s) Where `-' means that the time consumption has exceeded 15 hours.}
\label{tab:time}
\end{table}

\begin{figure}
    \centering
    \includegraphics[width=0.9\columnwidth]{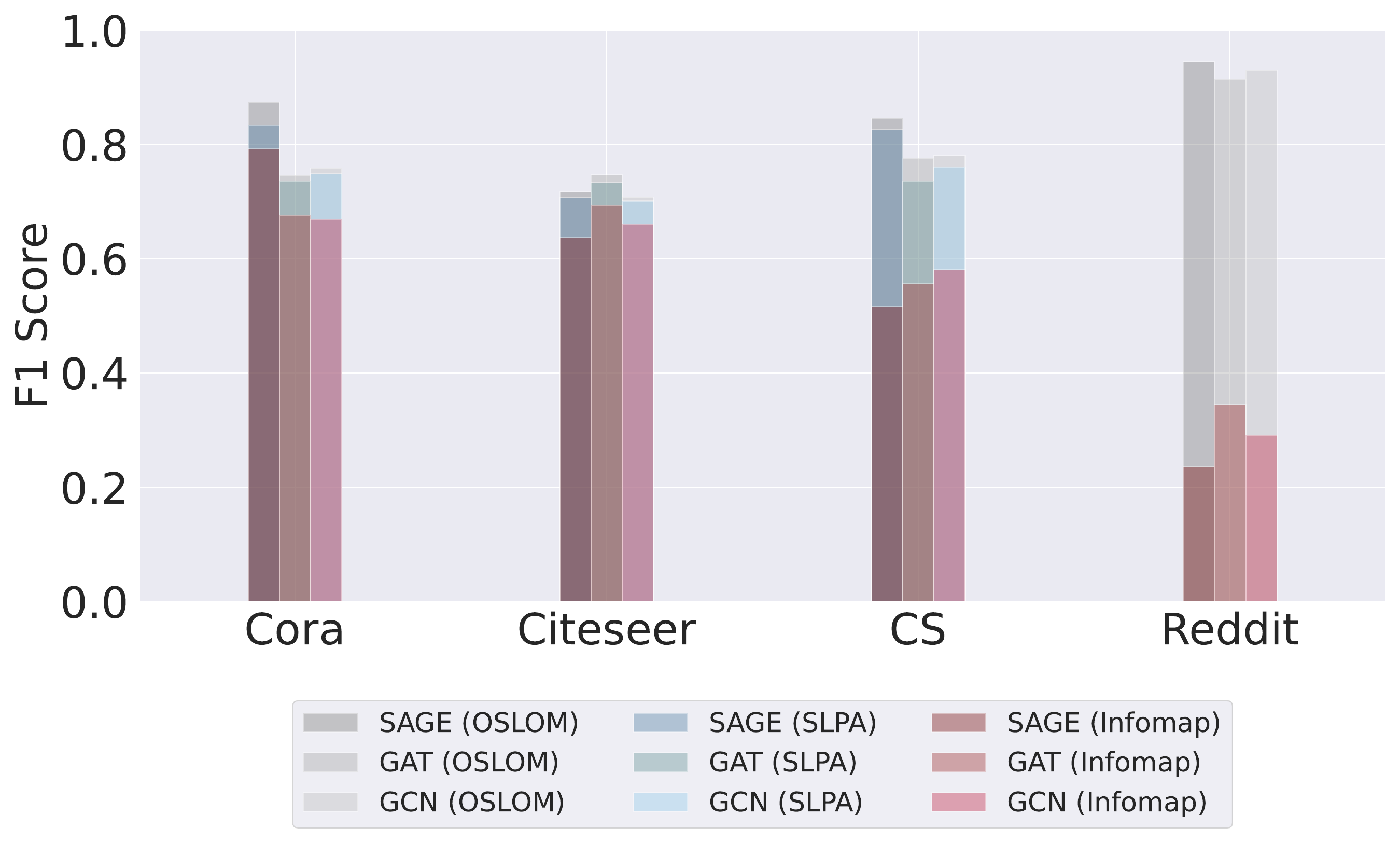}
    \caption{The comparison of Macro F1 scores of different community detection methods on four datasets and three GNN backbones.}
    \label{fig:cdf1}
\end{figure}

For the combination of Louvain and OSLOM coarse and fine granularity proposed in the main text, we give a total of two structures. For small datasets, Louvain's dataset delineation is used as the initialized community and OSLOM is used to optimize the community structure. For large datasets, we evaluate the conductivity of the initialized communities divided by Louvain, select the communities with poor conductance for fine-grained division, and optimize the community structure using OSLOM.

\begin{table*}[htbp]\small
\centering
\newcolumntype{Z}{>{\centering\let\newline\\\arraybackslash\hspace{0pt}}X}
\begin{tabularx}{\textwidth}{X|ZZ|ZZ|ZZ|ZZ}
\toprule[1.5pt]
 & Cora(M) & Cora(O) & Citeseer(M) & Citeseer(O) & CS(M) & CS(O) & Reddit(M) & Reddit(O) \\
\midrule
Entropy & 1.8197 & 1.8311 & 1.7434 & 1.7533 & 2.3961 & 2.4087 & 3.3967 & 3.4013 \\
IR & 0.2024 & 0.2200 & 0.3068 & 0.3766 & 0.0291 & 0.0285 & 0.0141 & 0.0116 \\
Gini & 0.2706 & 0.2587 & 0.1400 & 0.1338 & 0.4328 & 0.4225 & 0.4314 & 0.4222 \\
\bottomrule[1.5pt]
\end{tabularx}
\caption{Class distribution and imbalance measures for mapped and original graphs. (M) denotes the mapped graph generated by CGE and (O) denotes the original graph.}
\label{bias}
\end{table*}

Table \ref{tab:time} presents the partitioning durations of different community detection methods across four datasets, while Figure \ref{fig:cdf1} illustrates their Macro F1 scores on various GNN backbones for the same datasets. When considering both results, SLPA and OSLOM demonstrate high model utility; however, SLPA's time consumption for large datasets is prohibitively high, rendering it impractical. Although Infomap is efficient in partitioning, its predictive performance is insufficient. Despite the relatively lower partitioning efficiency of OSLOM, we elected to utilize this method because CGE only necessitates a single community detection process throughout the workflow. We prefer the method with superior performance rather than optimizing time. Moreover, the efficiency experiments presented in the main text confirm that, despite OSLOM's lower partitioning efficiency compared to other methods, it still offers significant advantages in model deployment compared to the baselines.

We carefully analyzed the characteristics of different community detection methods. The conventional OSLOM method is compatible with a wide range of graph structures, including directed graphs. We mitigated its high computational cost in large-scale networks by integrating coarse and fine-grained methods. In comparison, SLPA relies more on the randomness of initialization, while Infomap places an undue emphasis on random walks. To ensure the robustness of CGE, the more stable OSLOM method was selected.

It is worth mentioning that the GSMU paradigm and its CGE framework proposed in this paper are a novel universal and efficient graph unlearning strategy, and its great potential supports the integration of any more effective community detection method or graph mapping algorithm.

\section{E. Evaluation of Class Imbalance in CGE}

In the framework design, it was considered that some nodes within a single community may represent noise at the feature or label level relative to the entire collective. To mitigate the complete dominance of the majority class, an intuitive approach was proposed based on majority voting and Euclidean distance-based local proximity. This allows label selection to better align with the community's local structure, thereby minimizing the negative impact on supervised learning. 

To validate the approach, we conducted experiments involving the class distribution of nodes in the generated mapped graph and the original graph. The class imbalance was quantified using three mathematical measures: class distribution vector entropy, class imbalance ratio (IR), and the Gini coefficient. The following outlines the methods used for calculating these measures:

\begin{itemize}
    \item \textbf{Class Distribution Vector Entropy:} The entropy of the class distribution vector was calculated using the formula:
    \begin{equation}
        H = - \sum_{i=1}^{n} p_i \log(p_i),
    \end{equation}
    where \( p_i \) is the proportion of nodes belonging to class \( i \), and \( n \) is the number of distinct classes. This measure quantifies the uncertainty or disorder within the class distribution.
    
    \item \textbf{Class Imbalance Ratio (IR):} The class imbalance ratio is defined as the ratio between the size of the largest class and the size of the smallest class:
    \begin{equation}
        IR = \frac{\text{Size of Largest Class}}{\text{Size of Smallest Class}}.
    \end{equation}
    A higher ratio indicates greater imbalance in the class distribution.

    \item \textbf{Gini Coefficient:} The Gini coefficient is calculated as:
    \begin{equation}
        G = 1 - \sum_{i=1}^{n} p_i^2,
    \end{equation}
    where \( p_i \) is the proportion of nodes in class \( i \). This measure provides a summary of the inequality of the class distribution, with higher values indicating greater imbalance.
\end{itemize}

The results of the experiment are summarized in Table \ref{bias}, where (M) denotes the mapped graph generated by CGE and (O) denotes the original graph.

The results indicate that the mapped graph generated by CGE did not introduce additional data bias when compared to the original graph.

\end{document}